\pdfoutput=1
\documentclass{article}

\usepackage[final,sglblindworkshop]{neurips_2026}
\workshoptitle{Interpretability as a Science}

\usepackage[utf8]{inputenc}
\usepackage[T1]{fontenc}
\usepackage{courier}
\usepackage{hyperref}
\hypersetup{
  hidelinks,
  pdftitle={What MoE Routing Controls: Causal Interventions on Expert Trajectories},
  pdfauthor={Mukund Athmaram, Manoj Rao, Liang Ma},
  pdfsubject={NeurIPS 2026 Workshop: Interpretability as a Science}
}
\usepackage{url}
\usepackage{booktabs}
\usepackage{amsfonts}
\usepackage{amsmath}
\usepackage{amssymb}
\usepackage{nicefrac}
\usepackage{microtype}
\usepackage{xcolor}
\usepackage{graphicx}
\usepackage{CJKutf8}
\usepackage{float}
\usepackage{placeins}

\title{What MoE Routing Controls: \\ Causal Interventions on Expert Trajectories}

\author{%
  Mukund Athmaram$^{1,2}$\thanks{Correspondence to: \href{mailto:ma2388@cornell.edu}{\texttt{ma2388@cornell.edu}};
  Work conducted during an AMD internship.}\quad
  Manoj Rao$^{1}$\quad Liang Ma$^{1}$\\
  $^{1}$\,Advanced Micro Devices Inc. (AMD)\qquad $^{2}$\,Cornell University
}

\begin{document}
\maketitle

\begin{abstract}
In mixture-of-experts (MoE) language models, a token is routed to a sequence of experts across layers.
We investigate how forcing tokens onto chosen
expert paths influences model behavior. We build a 
target path from the experts that a concept's or role's tokens most often use, 
and we add a fixed bias to
each target expert's router logit during generation to steer the token toward this expert path. We study two kinds of targets. Format targets concern the form of the output, and concept targets concern its topic. In Qwen3-30B-A3B, we show that steering a format path changes the text the
model writes. We also show that steering a concept path shifts the model's next-token probabilities toward the
concept, even when the path is built in another language, yet this shift has little effect on the generated text. We investigate why formats reach the text but concepts do not by steering only part of the generation, and find
evidence that a format, once started, reinforces itself. We show that the concepts religion and money steer most strongly, and that the effect is concentrated in the middle layers.
 We attempt to explain why the strength of the effect varies by concept through an analysis of embedding
distinctiveness and word frequency.
\end{abstract}

\section{Introduction}
Mixture-of-experts (MoE) models send each token to a small subset of experts in every
layer~\cite{shazeer,switch,stmoe,mixtral}. At layer $\ell$, a linear router reads the token's hidden state
$h_\ell$ and gives each expert a score, $r_\ell = W_\ell^{g} h_\ell$. The top-$k$ experts are selected, and their
outputs are combined with softmax weights over their scores,
\begin{equation}
\mathcal{E}_\ell = \operatorname*{top\text{-}k}_{j}\, r_{\ell,j},
\qquad
g_{\ell,j} = \frac{e^{\,r_{\ell,j}}}{\sum_{j'\in\mathcal{E}_\ell}e^{\,r_{\ell,j'}}},
\qquad
y_\ell = \sum_{j\in\mathcal{E}_\ell} g_{\ell,j}\,f_{\ell,j}(h_\ell),
\label{eq:router}
\end{equation}
where $f_{\ell,j}$ is expert $j$'s feed-forward network. We call the selected sets $\mathcal{E}_\ell$ across
layers the token's expert path, and this path is what we intervene on. \citet{polysemantic} show that routing is
a linear read of the hidden state, and that single experts are polysemantic while multi-layer paths are
monosemantic, grouping tokens by semantic function across languages and surface forms. For example, they show that the token
``\texttt{:}'' follows different paths when it marks a type annotation, introduces a list, or separates hours
and minutes. Our format experiment in \S\ref{sec:format} uses this colon example from their work. We build on
this finding by forcing tokens onto these paths and measuring how the model's output
changes. \citet{polysemantic} suggest steering expert selection by perturbing the
hidden-state subspace that the router reads. We instead set the selection directly by biasing the router logits.
This changes only which experts run, while each expert receives the same input, so a change in the output can be
attributed to the choice of experts. It also places the exact target expert at each layer and lets us swap in
random or matched experts as controls. The cost is that a strong bias gives the target a larger gate weight than
it usually receives, which we examine in \S\ref{sec:dynamics}.

Earlier work has steered MoE router logits. \citet{steermoe} find experts linked to a behavior by comparing
paired safe and unsafe prompts. They then promote or suppress those experts through the router logits to control
behaviors such as safety and faithfulness. That method steers individual experts chosen for a target behavior. We use the same kind of router-logit intervention, but we steer a
full cross-layer path built from the experts that a concept's own tokens share. This lets us ask whether a path's
meaning transfers to the output, and we measure the effect on both the next-token distribution and the generated
text. Other work steers transformers by adding directions to the residual
stream~\cite{actadd,repe,arditi}, and this kind of steering often works for some behaviors and inputs but not
others~\cite{steervec}. \citet{rare} treat routing and behavioral steering as separable. They project
perturbations onto the router's null space so that steering leaves routing unchanged. We intervene on routing
itself and ask what that intervention controls on its own. How to interpret experts is also debated. Some work
finds that experts are fine-grained, interpretable specialists~\cite{expertstrikesback}. Other work argues that
apparent specialization reflects hidden-state geometry, and that routing topology can have little effect on
quality even when individual experts matter causally~\cite{myth,geomrouting}. Our experiments add evidence to this
debate by testing directly whether the experts a token is routed to change what the model outputs.

We study two kinds of paths. Format paths, such as the path of a colon in clock times, concern the form of the
output, while concept paths, such as those for religion or money, concern its topic. We show that forcing a format path
changes the text the model writes, while forcing a concept path shifts the model's next-token probabilities
toward the concept and leaves the generated text largely unchanged. We then study why format steering is reflected in the generated text while concept steering is not, by analyzing
self-reinforcement during generation. We also study which concepts can be steered, where in the network steering
acts, and how the results carry over to other MoE architectures.

\paragraph{Contributions.}
\begin{itemize}
\item A path-level steering method that forces tokens onto cross-layer expert paths built from a concept's or
role's own tokens, by biasing the router logits without editing the hidden state (\S\ref{sec:method}).
\item Evidence that format paths change the generated text. Forcing a colon in code onto the clock-time path
raises the digit fraction of the output to $44\times$ that of a random path (\S\ref{sec:format}).
\item Evidence that concept paths shift next-token probabilities across languages. A path built from Chinese
religion tokens raises the probability of English religion words by $7\times$ on the primary and $12\times$ on a
held-out vocabulary, with little change to the generated text. Steering only part of the generation suggests that
formats reach the text because they reinforce themselves (\S\ref{sec:concept}).
\item A consistent ordering of which concepts steer, with religion and money strongest, that is not explained by
vocabulary distinctiveness or word frequency (\S\ref{sec:law}).
\item Evidence that the identity of the promoted expert matters beyond its gate weight, activation frequency or
output norm, and that steering acts in the middle layers (\S\ref{sec:dynamics}).
\item A comparison across four MoE architectures, showing that path distinctiveness tracks routing width and that
cross-model comparisons need to hold the routed weight fixed (\S\ref{sec:general}).
\end{itemize}

\section{Method}
\label{sec:method}
\paragraph{Models.} Our main model is Qwen3-30B-A3B~\cite{qwen3}, the model in Figure~1 of
\citet{polysemantic}. It has 48 layers with 128 experts each and top-8 routing. We also test OLMoE, Qwen1.5-MoE and
DeepSeek-V2-Lite (\S\ref{sec:general}).

\paragraph{Building a target path.} We take the target path from the model itself. For a concept or role, we
choose a few exemplar tokens that clearly express it, such as the tokens ``prayer'', ``scripture'' and ``faith'' for religion. We run
each exemplar through the model and record its top-1 expert at every layer. The target expert at each layer is
the most common top-1 expert across the exemplars $\mathcal{W}_c$,
\begin{equation}
j^\star_\ell(c) = \operatorname*{arg\,max}_{j}\, \big|\{ w\in\mathcal{W}_c : \operatorname*{arg\,max}_{j'} r_{\ell,j'}(w) = j \}\big|,
\label{eq:target}
\end{equation}
so the path is the one that the concept's own tokens most often share.

\begin{figure}[b]
  \centering
  \includegraphics[width=0.82\linewidth]{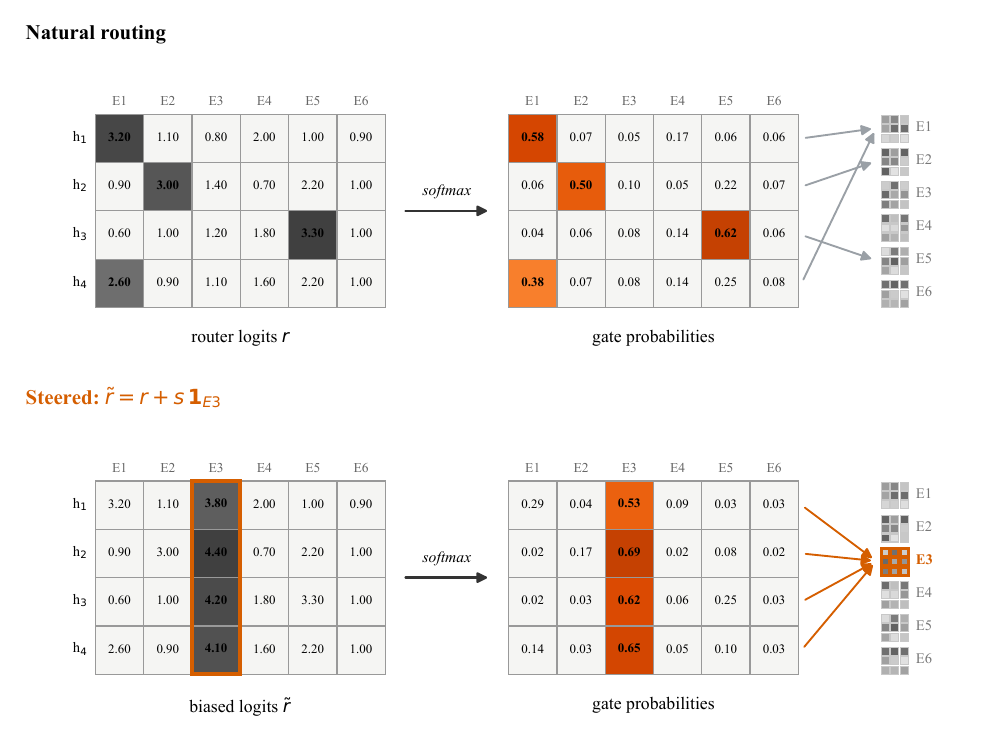}
  \caption{The steering intervention. Top: at a targeted layer, the router scores the experts and selects the
  top-$k$ (shaded), and we record the top-1. Bottom: we add a constant $s$ to the target expert's logit before
  selection, $\tilde{r} = r + s\,\mathbf{1}_{j^\star}$. The target joins the selected set with the other $k-1$
  experts and receives a larger gate weight. Only the routing logits change.}
  \label{fig:method}
\end{figure}

\paragraph{Steering.} To force a token onto the path, we add a constant bias $s$ to the target expert's router
logit at each steered layer, $\tilde{r}_\ell = r_\ell + s\cdot \mathbf{1}_{j^\star_\ell}$, and let top-$k$
selection proceed as usual (Figure~\ref{fig:method}). The target joins the selected set, and the router still
chooses the other $k-1$ experts. Unless stated otherwise, we steer the middle layers $\ell\in[16,31]$ at $s=8$,
where \S\ref{sec:dynamics} shows the effect is concentrated.

\paragraph{Measuring the effect.} A path could change what the model leans toward without changing what it
writes, so we measure both. The distribution readout is the next-token probability on the target's tokens, such
as digits or the first subtokens of a list of religion words. The text readout lets the model generate and
scores what it writes, such as the fraction of characters that are digits or the fraction of words that appear
in the concept's word list. We report each effect relative to the random path, with paired significance tests and
95\% bootstrap confidence intervals over prompts.

\paragraph{Controls.} Any large change to routing could disturb the output, so we compare each steered path with
a random path that receives the same bias, with one expert drawn uniformly at random per layer. To check that
a rise in a concept's words comes from that concept's own path, we also steer other concepts' paths and measure
every concept's words (\S\ref{sec:law}). A concept path could also raise its words simply by repeating the
tokens it was built from. Multilingual models share much of their concept information across languages~\cite{langneutral}, while
Chinese and English words share no tokens. We therefore build concept paths from Chinese tokens and measure
English words, and we
also measure a second, held-out word list that played no part in building the path.

\FloatBarrier
\section{Format steering changes the generated text}
\label{sec:format}
We begin with format, where the effect of a path is easy to see in the text. \citet{polysemantic} show that the
token ``\texttt{:}'' takes different paths when it marks a Python type annotation and when it separates hours and
minutes. We steer a type-annotation colon onto the clock-time path (layers 8--39) and let the model continue.
Without steering, the model writes Python types. We show that with steering, it writes digit strings
(Table~\ref{fig:gen}). The digit fraction of the output rises from 0.087 to 0.792, compared with 0.018 for the
random path, a $44\times$ increase over random. Output entropy stays between 2.3 and 2.7, so steering does not
collapse the output. The three uses of the colon take different experts at 26 of the 32 steered layers, and across
all six ordered pairs of uses, steering raises the target use's tokens to 2 to 40 times the random path. Forcing the path is therefore enough to change what the model
writes. Format is a property of the output's form, however, and the next question is whether a path can also
carry meaning.

\begin{table}[!htb]
  \caption{Format steering changes the generated text. Moving a type-annotation ``\texttt{:}'' onto the clock-time
  path turns type completions into digit strings. The digit fraction is 0.09 without steering, 0.79 with steering
  and 0.02 for the random path.}
  \label{fig:gen}
  \centering
  \begin{tabular}{lll}
    \toprule
    prompt & unsteered (writes a type) & steered to time path (writes digits) \\
    \midrule
    \texttt{values:} & \texttt{List[int] = [1, 2, 3} & \texttt{188108824081} \\
    \texttt{y:}      & \texttt{float = f(3.0)}       & \texttt{040610000061} \\
    \texttt{total:}  & \texttt{int = 0}              & \texttt{120,3,4,5,,8}  \\
    \texttt{result:} & \texttt{int = h(5)}           & \texttt{843.01 in,002} \\
    \bottomrule
  \end{tabular}
\end{table}

\FloatBarrier
\section{Concept steering across languages}
\label{sec:concept}
\begin{figure}[!htb]
  \centering
  \includegraphics[width=0.8\linewidth]{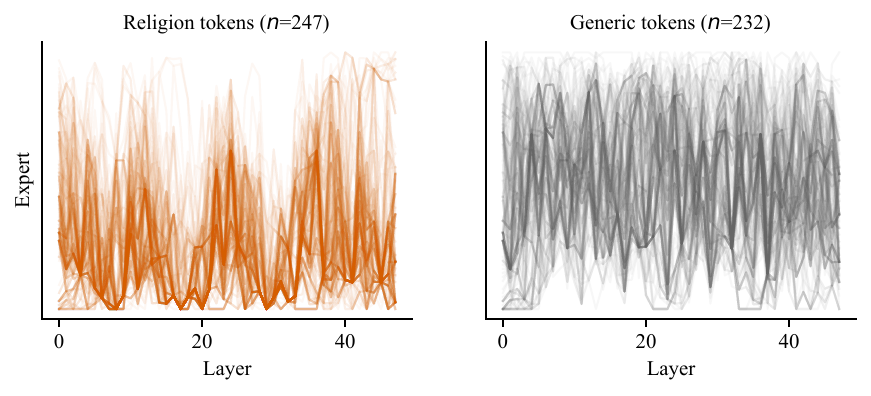}
  \caption{Top-1 expert trajectories for 247 religion-context positions (left) and 232 generic positions (right).
  We use the layered layout of \citet{polysemantic}, a Sugiyama-style layout~\cite{sugiyama} that reorders experts at
  each layer to reduce edge crossings, with one shared ordering for both panels.}
  \label{fig:relpath}
\end{figure}

\paragraph{Concept tokens share experts.} A concept path is only meaningful if the concept's tokens actually
share experts, and we show that they do. Religion tokens agree on the same expert at each layer far more often than generic tokens
(mean purity 0.47 against 0.19) and form a coherent band of shared experts (Figure~\ref{fig:relpath}).

\begin{figure}[!htb]
  \centering
  \includegraphics[width=0.82\linewidth]{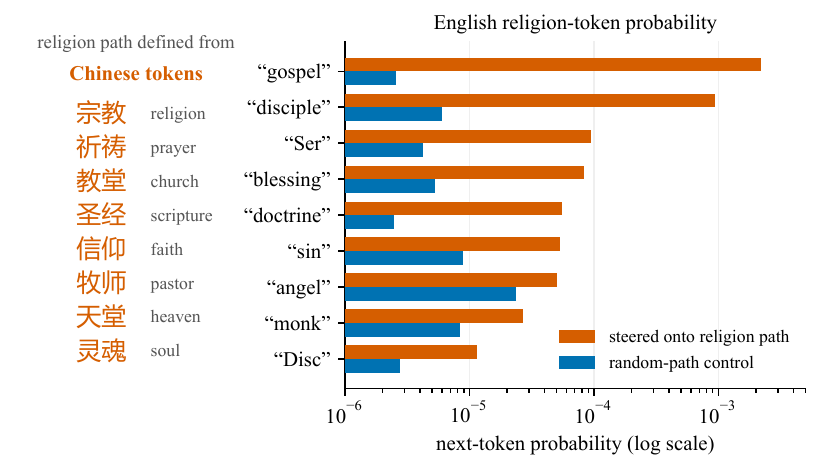}
  \caption{Cross-lingual concept steering. Left: Chinese exemplar tokens used to build the religion path (8 of 20
  shown). Right: probability of English religion tokens under the religion path and a random path (L16--31,
  $s=8$, 12 English prompts, averaged over 20 generated tokens).}
  \label{fig:cross}
\end{figure}

\paragraph{Steering a neutral prompt toward a concept.} We build the religion path from
\begin{CJK}{UTF8}{gbsn}Chinese\end{CJK} religion tokens and steer 60 neutral English prompts, such as
\texttt{"Let me tell you about"}, onto it. We show that this raises the probability of English religion words to
$7.1\times$ the random path, and to $12.4\times$ on the held-out word list. The effect is positive on 54 and 57 of the 60 prompts and exceeds each of 30 different random paths, at $6.2\times$ their mean. The rise is spread across many religion words, such as the tokens ``gospel'', ``disciple'' and
``blessing'' (Figure~\ref{fig:cross}). The money concept also steers, at $3.6\times$ the random path. On the held-out money list it stays at the
random level, which fits money words being common in neutral text and having little room to rise. Because the
path is built in Chinese and measured in English, these effects reflect the concept rather than repeated tokens.
The absolute probabilities stay small, rising from $1.2\times10^{-3}$ to $8.3\times10^{-3}$ for religion, which
raises the question of whether the shift reaches the text the model writes.

\begin{figure}[!htb]
  \centering
  \includegraphics[width=0.6\linewidth]{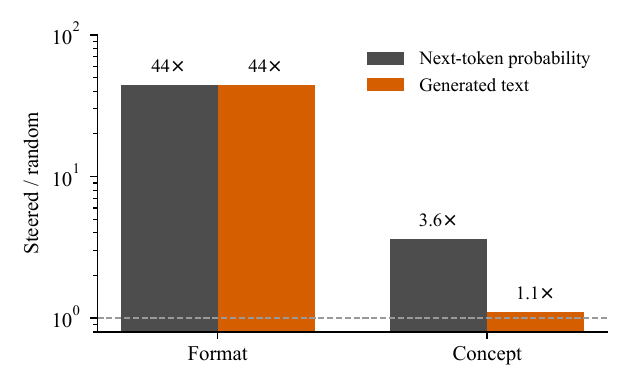}
  \caption{Steering effect on next-token probability and on generated text, relative to the random path (log
  scale; dashed line, no effect). Format (digits) changes both. Concept (money) changes probability but not text.}
  \label{fig:dissoc}
\end{figure}

\paragraph{Distribution and generated text.} For concepts, the two readouts disagree (Figure~\ref{fig:dissoc}).
The distribution leans toward money, with money-word probability at $3.6\times$ the random path, but the model
does not write money-themed text. The fraction of money words in the generated text stays near 0.01, no higher
than for the random path. For format, the two readouts agree, and the lean carries into the text. Routing therefore controls what the
model reaches for in both cases, but only the format change becomes what the model writes, and we investigate why below.

\begin{figure}[!htb]
  \centering
  \includegraphics[width=0.75\linewidth]{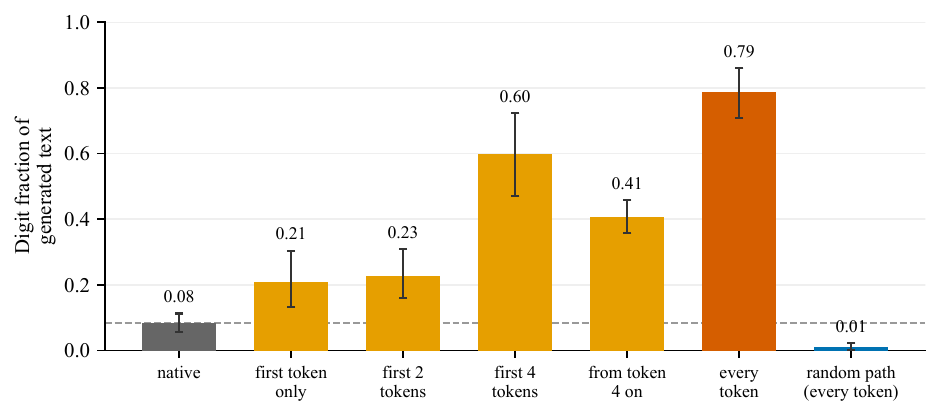}
  \caption{Digit fraction over 12 generated tokens when the clock-time path is promoted on different generation
  steps ($N{=}30$ prompts). The dashed line marks the native level, and error bars are 95\% prompt-bootstrap
  intervals.}
  \label{fig:pulse}
\end{figure}

\paragraph{Why format persists and concept does not.} Both paths change what the model leans toward, so the
difference must arise during generation. Once the model writes a digit, that digit becomes context and makes more
digits likely. If this self-reinforcement drives the format effect, steering only the first few tokens should
already start a digit run. We show it does (Figure~\ref{fig:pulse}). Steering only the first token raises the digit
fraction from 0.085 to 0.21, and steering the first four raises it to 0.60, against 0.79 for steering every token. Steering that starts only after the fourth token still reaches 0.41, so
the effect does not depend on the first position either. We hypothesize that concept words do not reinforce themselves in this way, so the text stays close to the native
continuation even though the push toward the concept is present at every step.

\FloatBarrier
\section{Which concepts can be steered}
\label{sec:law}
\begin{figure}[!htb]
  \centering
  \includegraphics[width=0.65\linewidth]{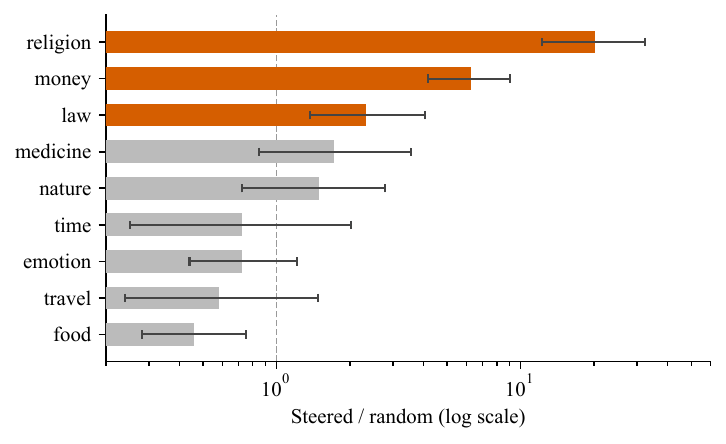}
  \caption{Steerability by concept. Steered-over-random concept-word probability ($N=60$ paired; log scale; $95\%$
  bootstrap CIs; a star marks $p<0.05$ with an interval above 1). Religion and money steer strongly and law weakly
  (red).}
  \label{fig:domains}
\end{figure}

Religion and money both steer, which raises the question of whether every concept does. We repeat the experiment
for nine concepts (Figure~\ref{fig:domains}). We show that religion and money steer strongly ($20\times$ and
$6.3\times$ the random path) and law steers weakly ($2.4\times$, with no effect on its held-out word list), while medicine, nature,
food, travel, emotion and time do not reliably steer. This ordering holds across our experiments.

\begin{figure}[!htb]
  \centering
  \includegraphics[width=0.75\linewidth]{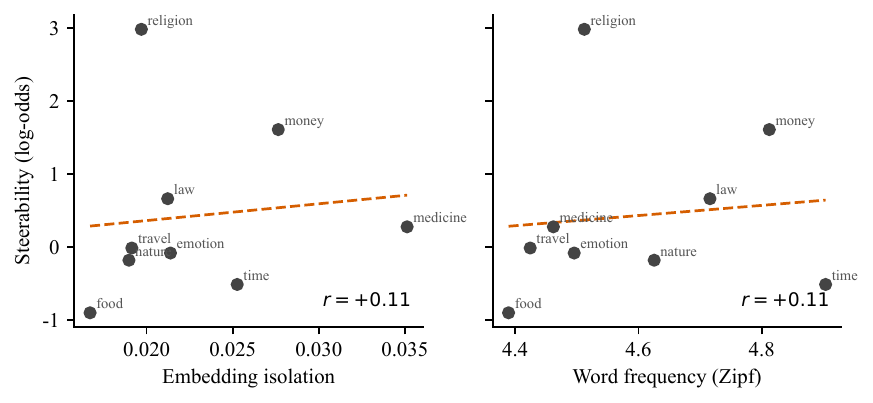}
  \caption{Steerability against embedding-space isolation (left) and lexical frequency (right). Pearson
  $r=+0.11$ for each, $n=9$.}
  \label{fig:refute}
\end{figure}

\paragraph{Explaining the ordering.} A natural explanation is that concepts with more distinctive or rarer
words are easier to steer, and we test both explanations (Figure~\ref{fig:refute}). We measure distinctiveness as the mean
within-concept cosine minus the mean between-concept cosine over 765 word vectors in English, Chinese and
Spanish. We pre-registered that two concepts built to be distinctive (religion, astronomy) would score above two
built to be generic (object, action), and they score below. In middle-layer hidden states, which are dominated by a single direction, any correlation
disappears after the standard all-but-the-top correction. We measure rarity by corpus frequency and by subtoken count, which agree closely. Neither measure predicts steerability. Religion, the strongest concept, even
has a higher baseline probability than most other concepts, which is the opposite of what a rarity account predicts. Language behaves differently from topic. Only 3 of 16
middle layers separate languages, and language cannot be steered, which fits the finding of \citet{polysemantic}
that the router does not read language, although this measurement uses a single token position.

Clean routing is not required either. Money's words agree on the same
expert only $48\%$ of the time, yet money steers strongly. Steering only the few experts that a concept's Chinese and English words share gives a much weaker effect (2 to 4 times) than steering the whole middle band
(17 to 28 times), so the effect comes from many layers together.

\begin{figure}[!htb]
  \centering
  \includegraphics[width=0.58\linewidth]{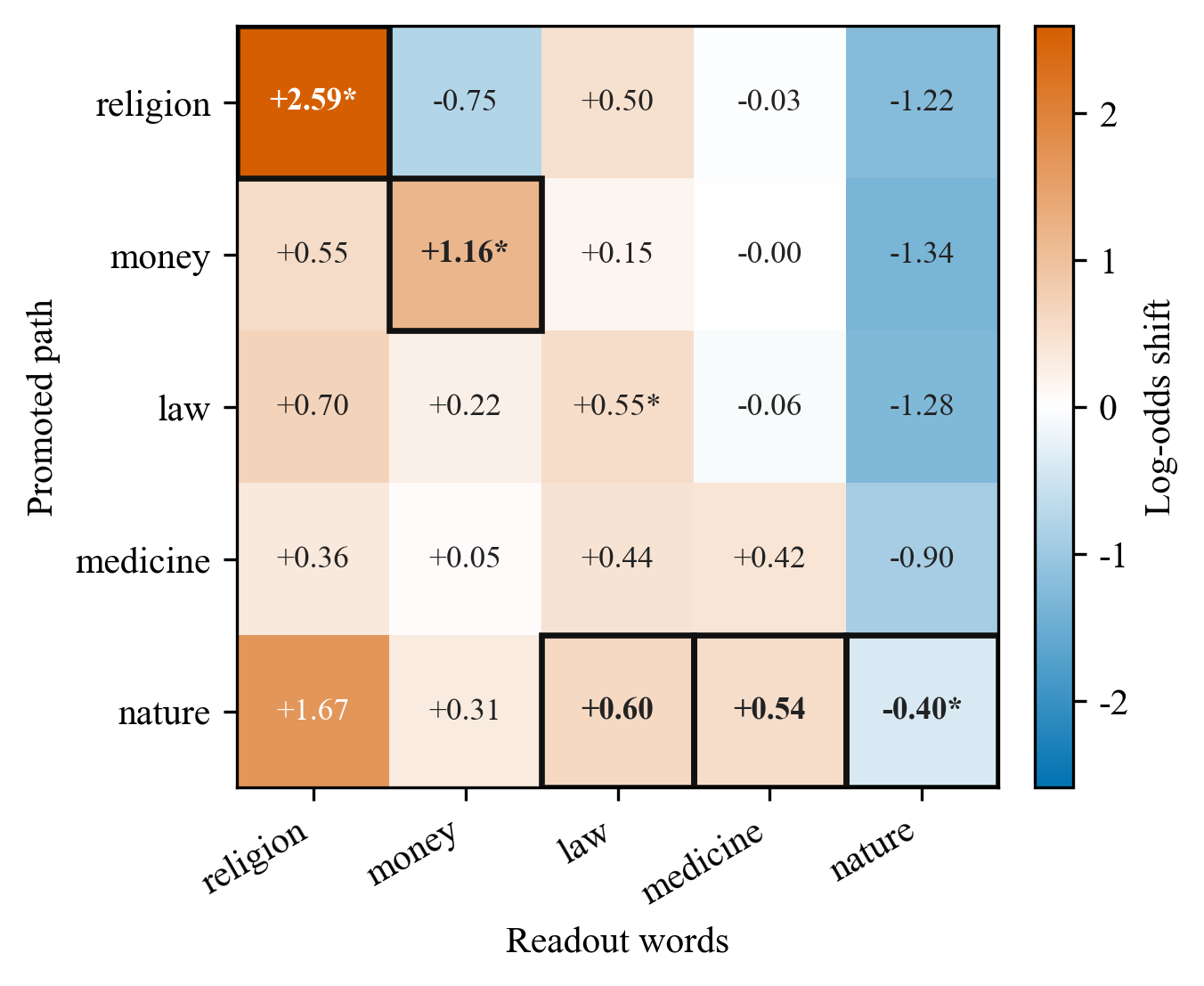}
  \caption{Concept specificity. Each cell is the paired log-odds shift over the random path when a promoted path
  (row) is read out on a concept's words (column). The boxed cell is each column's maximum ($N{=}60$ paired; a
  star marks $p<0.05$).}
  \label{fig:specificity}
\end{figure}

\paragraph{Specificity.} A concept that steers should raise its own words more than other concepts' words. We
promote each concept's path and measure every concept's words (Figure~\ref{fig:specificity}). We show that the
two concepts that steer strongly are specific. Promoting religion raises religion words most ($+2.59$ log-odds), and promoting money raises money words most ($+1.16$). Weaker concepts are less
specific. Promoting nature raises religion words ($+1.67$) and lowers nature words ($-0.40$), and promoting law raises
religion words ($+0.70$) about as much as law words ($+0.55$). Overlap between paths does not explain this
leakage, since path agreement between concepts does not predict it. Specificity therefore
follows the same ordering as steerability.

\FloatBarrier
\section{Where steering acts and what drives it}
\label{sec:dynamics}
\begin{figure}[!htb]
  \centering
  \includegraphics[width=0.75\linewidth]{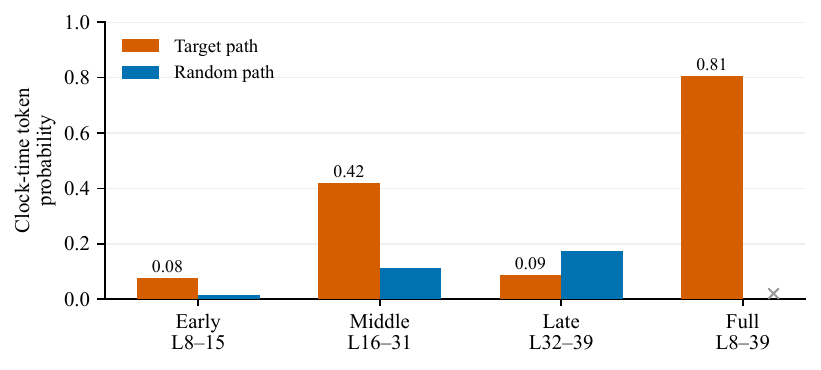}
  \caption{Clock-time token probability by steered layer band ($s=16$, $N=20$). No random path was run for the full band (x).}
  \label{fig:bands}
\end{figure}

\begin{figure}[!htb]
  \centering
  \includegraphics[width=0.8\linewidth]{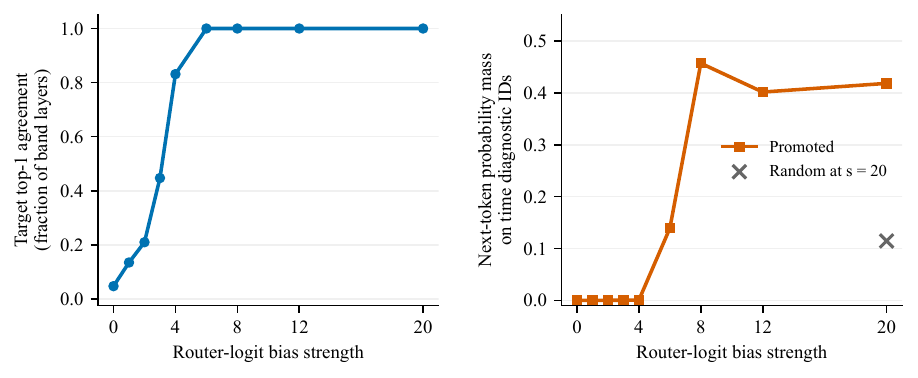}
  \caption{Target top-1 agreement (left) and time-diagnostic probability (right) across strengths (L16--31,
  $N=20$).}
  \label{fig:dose}
\end{figure}

\paragraph{Layers and strength.} We next ask where in the network the path matters. For format, steering the
middle layers raises the clock-time probability to 0.418, against 0.076 for early and 0.088 for late layers
(Figure~\ref{fig:bands}). For religion, the two middle eight-layer windows give $3.5\times$ and $8.4\times$ the
random path, and the other windows give no effect. Because these windows have equal width, the middle-layer effect is not due to the middle band being wider. This matches causal tracing, which places factual recall in middle-layer feed-forward modules~\cite{rome}, and \citet{steermoe}, who find behavior-linked
experts in middle layers. Strength matters as well. At $s=6$ the target is already the top-1 expert at every
steered layer, yet raising $s$ to 8 still increases the clock-time probability from 0.139 to 0.457
(Figure~\ref{fig:dose}). The bias therefore does more than select the target. It also raises the target's gate weight.

\begin{figure}[!htb]
  \centering
  \includegraphics[width=0.65\linewidth]{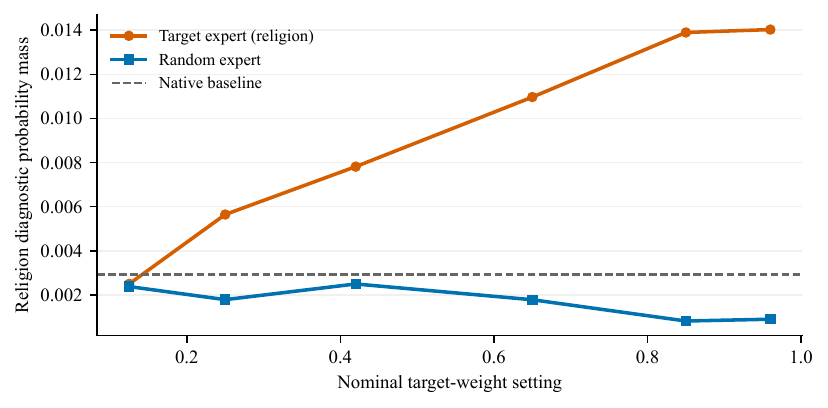}
  \caption{Religion-word probability when the religion expert or a random expert is forced into the selected set
  at a range of weights ($N{=}60$ paired). The dashed line is the native baseline. Weights are set before
  renormalization.}
  \label{fig:factorial}
\end{figure}

\paragraph{Expert identity and gate weight.} This raises the question of whether the effect comes from the target expert or from giving any expert a large
weight. The two cannot be fully separated, because any bias large enough to select an expert also raises its gate
weight. A naturally top-1 expert carries a gate weight of about 0.42, and our bias raises the target's to about
0.96. The model does not select experts at near-zero weight, so we compare identities across a range of
realistic weights. We force either the religion expert or a random expert
into the selected set at a range of weights (Figure~\ref{fig:factorial}). We show that religion probability rises
with weight for the religion expert, from $0.0025$ to $0.014$ against a native baseline of $0.0029$, while the random expert stays at or below the native baseline. At a weight of 0.42, close to that of a natural
top-1 expert, the religion expert still raises religion probability to $3.45\times$ native, and at
the equal-share weight of 0.125 the effect disappears ($0.95\times$). We also show that random experts matched to the religion expert on activation frequency or
output norm do not reproduce the effect (Table~\ref{tab:confound}). The identity of the expert sets the
direction of the effect, and its weight sets the size.

\begin{table}[!htb]
  \caption{Religion mass of each steered condition relative to its comparison. Weights are the target's nominal
  weight, with the delivered weight in parentheses.}
  \label{tab:confound}
  \centering
  \small
  \begin{tabular}{llcc}
    \toprule
    steered condition & compared with & primary & held-out \\
    \midrule
    weight 0.42 (0.55) & native & $3.45\times$ $[1.44, 9.82]$, $p=0.003$ & $2.91\times$ $[0.76, 5.52]$, n.s. \\
    weight 0.42 (0.55) & random path & $3.91\times$, $p=0.04$ & -- \\
    weight 0.125 (0.27) & native & $0.95\times$ $[0.49, 1.81]$ & -- \\
    random path & native & $0.88\times$, n.s. & -- \\
    full steering & frequency-matched & $3.7\times$, $p<10^{-4}$ & $7.4\times$, $p<10^{-4}$ \\
    full steering & norm-matched & $6.6\times$, $p<10^{-4}$ & $1.5\times$, $p=0.005$ \\
    \bottomrule
  \end{tabular}
\end{table}

\FloatBarrier
\section{Other MoE architectures}
\label{sec:general}
Finally, we ask whether these results carry over to other MoEs. A concept can only be steered if it has its own
path, so we first check path distinctiveness on OLMoE~\cite{olmoe}, Qwen1.5-MoE~\cite{qwen15moe} and
DeepSeek-V2-Lite~\cite{deepseekv2}. The top-8 models give each concept its own path, while the top-4 and top-6 models send several
concepts through one shared path (Table~\ref{tab:general}). Steering follows this pattern, from $3.2\times$ for
religion on Qwen1.5-MoE to no effect on DeepSeek. We think DeepSeek's fine-grained routing with always-on shared
experts leaves no single expert whose promotion concentrates a concept. The four models differ in many ways besides routing width, so we
report this as a correlation. 

\begin{table}[!htb]
  \caption{Path distinctiveness and religion steering across four MoEs. \emph{Distinct paths}: concepts (of six)
  with their own path. \emph{Winner vote}: mean fraction of a concept's words that agree on the top expert per
  band layer. \emph{Religion steering}: steered over random, 60 prompts.}
  \label{tab:general}
  \centering
  \begin{tabular}{llccc}
    \toprule
    & & distinct & winner & religion \\
    model & routing & paths & vote & steering \\
    \midrule
    Qwen3-30B-A3B (ref.) & top-8, 128 experts & 6/6 & 0.64 & 7.1$\times$ \\
    OLMoE-1B-7B        & top-8, 64 experts  & 6/6 & 0.74 & 1.3$\times$ \\
    Qwen1.5-MoE-A2.7B    & top-4, 60 experts  & 3/6 & 0.86 & 3.2$\times$ \\
    DeepSeek-V2-Lite     & top-6, 64 experts  & 5/6 & 0.83 & 0.2$\times$ \\
    \bottomrule
  \end{tabular}
\end{table}

\paragraph{Keeping the routed weight fixed.} Steering magnitudes also depend on how a model weights its selected
experts. Qwen rescales the selected weights to sum to one, while OLMoE does not, so the same bias raises OLMoE's
total routed weight from about 0.4 to about 0.98. With the original bias, format steering does not transfer to
OLMoE (Table~\ref{tab:olmoe}). When we rescale the steered weights to keep OLMoE's native total, the digit
fraction rises from 0.043 to 0.281, an increase of 0.238, although only 4 of 40 outputs begin with a valid time. Concept steering raises
religion probability above the random path but not above the native model, as on Qwen. Preserving the weight also makes the steered text more likely under the unmodified model,
lowering the negative log-likelihood of format continuations from 8.6 to 5.9 nats per token, though this remains
well above the native 0.8. Comparisons of steering
across models therefore need to hold the routed weight fixed.

\begin{table}[!htb]
  \caption{OLMoE results under the original promotion and with the native routed weight preserved. Religion
  probability is the mean diagnostic probability on the primary word list ($\times10^{-3}$).}
  \label{tab:olmoe}
  \centering
  \begin{tabular}{lccccc}
    \toprule
    & & \multicolumn{2}{c}{original} & \multicolumn{2}{c}{weight preserved} \\
    \cmidrule(lr){3-4}\cmidrule(lr){5-6}
    & native & target & random & target & random \\
    \midrule
    digit fraction of output       & 0.04 & 0.04 & 0.00 & \textbf{0.28} & 0.00 \\
    religion probability           & 3.2  & 4.1  & 3.8  & 3.0  & 2.3 \\
    outputs with a religion word   & 5/60 & 1/60 & 1/60 & 1/60 & 1/60 \\
    \bottomrule
  \end{tabular}
\end{table}

\FloatBarrier
\section{Removing an expert during natural use}
\label{sec:suppression}
Our experiments force paths onto tokens. A complementary question is whether the experts on a path matter when
the model selects them itself. As a first test, we remove one naturally selected target expert at one layer and
compare with removing a matched non-target expert, on 64 held-out cloze prompts per concept
(Table~\ref{tab:suppression}). We measure the effect in two ways. Diagnostic probability is the
next-token probability on the concept's tokens, such as religion words or clock-time digits. Reference
log-probability is the log-probability the model assigns to the expected answer for the prompt. A selective
impairment would lower both more for the target expert than for the control. Removing a single expert does not produce a reliable selective drop, although for time the native probability is already 99.98\%, which leaves little room to fall. This result may mean
that the model compensates for one missing expert. Testing whole paths is a natural next step.

\begin{table}[!htb]
  \caption{Removing one naturally selected expert: target minus matched control, with 95\% CIs. Diagnostic
  probability is in percentage points.}
  \label{tab:suppression}
  \centering
  \begin{tabular}{llll}
    \toprule
    concept & prompts & measure & target $-$ control \\
    \midrule
    religion & 42 (norm-matched) & diagnostic probability & $-0.09$ $[-0.43, 0.24]$ \\
    religion & 42 (norm-matched) & reference log-probability & $-0.04$ $[-0.11, 0.02]$ \\
    religion & 64 & diagnostic probability & $-0.26$ $[-0.62, 0.04]$ \\
    time & 64 & reference log-probability & $\phantom{-}0.02$ $[-0.03, 0.08]$ \\
    \bottomrule
  \end{tabular}
\end{table}

\FloatBarrier
\section{Limitations and future work}
Our results show that a forced trajectory shifts the output in the tested settings. Our main results come from
one model, Qwen3-30B-A3B, and the other architectures test path structure and a subset of the steering results.
The bias is also a strong intervention, since at $s=8$ across 16 layers it gives the target expert a larger gate weight than
it usually receives, and Section~\ref{sec:dynamics} shows that this weight affects the size of the effect. Whether routing also mediates
the model's natural semantic computation is an open question. Removing a single expert during natural use was inconclusive (\S\ref{sec:suppression}), and testing whole-path necessity remains future work. In the weight experiment, we set the target's weight before the selected weights are rescaled, so the
weight the religion and random experts actually receive is not exactly the same. The four models we compare also
differ in size, training data and expert design as well as in how many experts each token uses, so routing width
is only one possible explanation for the differences between them. Explaining which concepts steer is also open,
since distinctiveness and frequency do not account for it (\S\ref{sec:law}).

\FloatBarrier
\section{Conclusion}
We show that promoting a data-derived target-expert trajectory causally changes an MoE model's behavior. Format paths change
the generated text, and concept paths shift the next-token distribution toward the target concept, even when the
path is built in another language. In our experiments, the difference follows whether the target reinforces
itself during generation, which points to self-reinforcement as the condition for routing control to persist in
the output. Looking more closely at concepts, we show that religion and money steer most strongly, an
ordering that neither distinctiveness nor word frequency explains, that the identity of the promoted expert
matters beyond its gate weight, and that steering acts in the middle layers. Across architectures, steering
depends on whether a concept has its own path. These results suggest that expert paths could serve as a simple
handle on the form of an MoE model's output and as a tool for studying how these models organize concepts.

\FloatBarrier

\bibliographystyle{plainnat}
\bibliography{refs}

\end{document}